\documentclass[acmsmall]{acmart}

\def\BibTeX{{\rm B\kern-.05em{\sc i\kern-.025em b}\kern-.08emT\kern-.1667em\lower.7ex\hbox{E}\kern-.125emX}}
    
\setcopyright{acmcopyright}
\acmJournal{TECS}
\acmYear{2021}\acmVolume{37}\acmNumber{4}\acmArticle{111}\acmMonth{7}
\acmDOI{10.1145/1122445.1122456}

\begin{document}


%
\title[PhiNets]{PhiNets: a scalable backbone for low-power AI at the edge}


%
\author{Francesco Paissan}
\authornote{Both authors contributed equally to this research.}
\email{fpaissan@fbk.eu}
\orcid{0000-0002-5553-7935}
\author{Alberto Ancilotto}
\authornotemark[1]
\email{aancilotto@fbk.eu}
\author{Elisabetta Farella}
\email{efarella@fbk.eu}
\affiliation{%
  \institution{E3DA Unit, Digital Society Center - Fondazione Bruno Kessler (FBK)}
  \streetaddress{Via Sommarive, 18}
  \city{Povo, Trento}
  \state{Italy}
  \postcode{38123}
}

%
\renewcommand{\shortauthors}{Paissan and Ancilotto, et al.}

%
\begin{abstract}
In the Internet of Things era, where we see many interconnected and heterogeneous mobile and fixed smart devices, distributing the intelligence from the cloud to the edge has become a necessity. Due to limited computational and communication capabilities, low memory and limited energy budget, bringing artificial intelligence algorithms to peripheral devices, such as the end-nodes of a sensor network, is a challenging task and requires the design of innovative methods. 
In this work, we present \textit{PhiNets}, a new scalable backbone optimized for deep-learning-based image processing on resource-constrained platforms. \textit{PhiNets} are based on inverted residual blocks specifically designed to decouple the computational cost, working memory, and parameter memory, thus exploiting all the available resources.
With a YoloV2 detection head and Simple Online and Realtime Tracking, the proposed architecture has achieved the  state-of-the-art results in (i) detection on the COCO and VOC2012 benchmarks, and (ii) tracking on the MOT15 benchmark. \textit{PhiNets} reduce the parameter count of 87\% to 93\% with respect to previous state-of-the-art models (EfficientNetv1, MobileNetv2) and achieve better performance with lower computational cost.
Moreover, we demonstrate our approach on a prototype node based on a STM32H743 microcontroller (MCU) with 2MB of internal Flash and 1MB of RAM and achieve power requirements in the order of 10 mW.
The code for the \textit{PhiNets} is publicly available on GitHub\footnote{\url{https://github.com/fpaissan/phinet_pl}}.
\end{abstract}

\keywords{Multi-Object Tracking, Neural Networks, Edge AI, Tiny ML}

%

%
\maketitle

\section{Introduction}
Over the past decade, we have witnessed two parallel trends. On one side, the increasing popularity of the internet of things, i.e., intelligent networked things everywhere, a consequence of the growing capabilities of the embedded systems, enhanced with capable processing units working at always increasing frequencies and offering attractive low-power modes \cite{cortex-m, raspberry, friendlyelec}. 
On the other side, with the advent of deep learning techniques, machine learning algorithms' size grows exponentially, thanks to the improvements in processor speeds and the availability of large training data. However, 
embedded systems cannot sustain the resource requirements of standard deep learning techniques, adequate for GP-GPUs \cite{Cerutti2019, Rusci2018, hao2021enabling}.

How then to compose the need for exploiting the opportunity to bring intelligence at the edge with the complexity of deep learning?
In this context, the junction point is TinyML \cite{wang2020convergence, xu2020edge}, a cutting-edge field that brings the machine learning (ML) transformative power to the performance- and power-constrained domain of tiny devices and embedded systems.

Among the several application domains in which exploring this novel trend, computer vision is one of the most popular. In this domain, object detection and tracking should be real-time, reliable, and accurate. Recent best-performing pipelines for multi-object detection and tracking imply using many computational resources, thus substantially limiting the application scenarios in which such techniques can be exploited. The current limitations mainly depend on the high computational cost of state-of-the-art techniques, which require GPUs for real-time inference and thus are not a good candidate for resource-constrained devices.

The most efficient solutions for Multi-Object Tracking (MOT) follows the tracking-by-detection paradigm. The tracking algorithm consists of an association algorithm based on the detected bounding boxes. Many pipelines are available for object detection. In particular, the main difference for what concerns object detection is the number of stages required to detect objects in one frame. The detection pipeline can be one-stage (single-shot bounding box regression) \cite{bochkovskiy2020yolov4, liu2016ssd} or two-stage (region proposal, object identification) \cite{he2017mask, ren2015faster}. The one-stage detectors are the most efficient from the computational complexity perspective; thus, they are the go-to solution for lightweight, real-time object detection.

The most popular one-stage detectors are YOLO and SSD. The core of the detection pipeline is the convolutional backbone for latent space representation of the images, which accounts for most of the Multiply-Accumulate operations (MACC). After that, the detection head is applied to the output of the backbone and outputs the bounding box regression.
Since the most expensive part of the pipeline is the backbone, many works have proposed scalable architectures to improve the efficiency in image processing \cite{howard2017mobilenets, tan2021efficientnetv2}. The ability of the networks to change computational requirements is an asset for embedded inferencing since the embedded platforms are diverse in hardware constraints. For example, typical IoT-oriented MCUs, such as the STM32F7 MCU or STM32L4, only have 320kB SRAM/1MB Flash and 32KB SRAM/256KB Flash, respectively, thus they cannot run the same networks. Also, developing a CNN architecture to tackle vision problems efficiently is not a new problem. Networks using minimal resources, such as MobileNets \cite{howard2017mobilenets} and EfficientNets \cite{tan2019efficientnet}, have already been optimized to allow state-of-the-art performance with less than $1B$ Multiply and Accumulate operations (MACCs), but with low performance on MCU scale computational resources.

Our work contributes to the state-of-the-art by proposing a novel scalable backbone,  \textit{PhiNets}, for detection and multi-object tracking on resource-constrained platforms. 
We prove the efficiency of \textit{PhiNets} by comparing them with existing lightweight backbones within a YOLOv2 \cite{redmon2017yolo9000} detection head and Simple Online Real-time Tracking (SORT) tracker \cite{bewley2016simple}.
Moreover, we implemented the tracking inference on off-the-shelf MCUs with state-of-the-art consumption (1.3mJ per frame).
Thus, our work has a meaningful impact in the fields of:
\begin{itemize}
\item embedded vision processing by proposing a new architecture family, \textit{PhiNets}, which pushes forward the state-of-the-art in object detection and  on tiny devices;
\item low-power image processing, since our pipeline requires only 1.3mJ per frame or 13mW at 10 fps;
\end{itemize}
\section{Related works}

\subsection{Scalable backbones}
In our work, we propose an MCU-friendly MOT pipeline based on deep learning. For this, we benchmarked many state-of-the-art approaches.
Our focus is on optimising the backbone, which is the initial part of the neural network in charge of learning and extracting the features needed to localize, classify, detect, and track objects. 
We focused on the backbone optimization since it accounts for the most considerable computational cost in the network inference. Thus, a reduction in the backbone complexity has a significant impact on the computational cost of the network.
Scalable backbones are an excellent solution for embedded processing since MCUs vary in resource availability, and different network architectures can be implemented based on the specific application domains.
 In the MobileNets paper, \cite{howard2017mobilenets, sandler2018mobilenetv2}, the authors presented a scalable backbone with good performance in vision benchmarks. In Howard et al. \cite{howard2017mobilenets}, the architecture is based on depth-wise separable convolutions and the authors presents a width and resolution multiplier. Both parameters are constrained in $[0, 1]$ (where 1 represents the standard architecture) and reduce parameter count and computational complexity quadratically. Instead, in Sandler et al. \cite{sandler2018mobilenetv2} inverted residual structures and linear bottlenecks represent the building blocks of the architecture. Moreover, the same scaling model presented by Howard et al. \cite{howard2017mobilenets} is applied.
Despite the model's ability to scale, the authors did not study any particular scaling model and its relationships with the model's performance in different vision tasks.
In the first EfficientNet paper \cite{tan2019efficientnet}, the authors propose a Neural Architecture Search (NAS) based approach for vision tasks that is also capable of scaling in depth. Moreover, they proved that scaling the neural network architecture one dimension at a time is less efficient than scaling all three dimensions (resolution, width, depth) simultaneously. The well-described compound scaling model, which consists of scaling all three dimensions by a power of the initial coefficients, proved to give a state-of-the-art performance in neural network scaling.
In EfficientNetV2 \cite{tan2021efficientnetv2}, the authors improved the model proposed by Tan et al. \cite{tan2019efficientnet} with training-aware NAS to optimise the model's parameter efficiency.

\subsection{Detection methods}
Object detection is a field of computer vision dealing with the detection of semantic objects in images. Different techniques have been developed in the past decades and will be hereafter compared based on the hardware implementation feasibility and computational complexity.
We can split the main detectors based on how many stages are required to extract the bounding boxes. The two-stage detectors, as for example Faster R-CNN \cite{ren2015faster} or Mask R-CNN \cite{he2017mask} are based on region proposal techniques that are then processed (i) to generate a region of interest or (ii) to perform object classification and bounding box regression. On the other hand, single-stage detectors such as YOLO (You Only Look Once) \cite{redmon2017yolo9000} or SSD (Single Shot Multibox Detectors) \cite{liu2016ssd} solve detection as a regression problem by learning to infer class probability and bounding box coordinates from input images. While two-stage detectors usually have higher accuracy scores with respect to single-stage detectors, they require more computational power and energy to infer a frame. Thus, to address MCU-friendly MOT applications, single-stage detectors are preferable. 


\subsection{Tracking methods}

Many object association techniques can be implemented for MOT, ranging from Intersection over Union (IoU) comparison to unified detection and tracking techniques \cite{zhang2020fairmot}. As it was for detection, there is a trade-off between computational complexity and performance, compared in terms of ID switches and MOTA score.
There are two categories of tracking algorithms, one in which algorithms perform tracking after a detection pipeline (e.g. SORT \cite{bewley2016simple}, DeepSORT \cite{wojke2017simple}, IoU) and the other category composed by algorithms that perform detection and tracking together (e.g. FairMOT \cite{zhang2020fairmot} and FairMOT Lite).

\begin{table}[ht]
\begin{tabular}{lcccc}
\hline
\multicolumn{1}{c}{} & IDs & MOTA \% & MOTP  & Hz (fps) \\ \hline
IoU                  & 287 & 61.6    & 0.116 & 5.33     \\ \hline
SORT                 & 48  & 54.3    & 0.172 & 5.31     \\ \hline
DeepSORT             & 47  & 55.9    & 0.175 & 3.64     \\ \hline
FairMOT              & 49  & 48.9    & 0.192 & 0.2      \\ \hline
FairMOT Lite         & 60  & 46.9    & 0.196 & 3.85     \\ \hline
\end{tabular}
\caption{Results of trackers on a sequence from MOT significant to smart cities environment. IoU, SORT, DeepSORT, and FairMOT lite use YOLOv5S as object detector.
The benchmarking is performed on a Intel(R) Core(TM) i9-10900KF CPU @ 3.70GHz}
\label{table:trackers}
\end{table}

Simple Online and Realtime Tracking (SORT) exploits the Hungarian association algorithm to associate bounding boxes from consecutive frames by maximizing the IoU score. The same approach is performed by Deep SORT, in which the score to be minimised is the distance (e.g. euclidean, cosine, correlation, etc.) between the latent representation of the content of the bounding boxes. This enables the algorithm to be more robust with respect to IoU because also visual attributes of the images are taken into account.

FairMOT instead is based on CenterNet \cite{duan2019centernet} and performs the detection and tracking together. This approach does not require anchors for the detection. In fact, the bounding box shape is inferred from the center of the blob. Some modification of this algorithm in which YOLOv5S is implemented in place of CenterNet is referred to as FairMOT Light here.

In Table \ref{table:trackers}, we quantitatively compared the presented trackers to help understanding trackers performance and computational cost (expressed in terms of frames per second). A more detailed discussion on the trackers is performed in Section \ref{sec:dettrack}.

\subsection{Vision-based MCU applications}
Although tiny vision is an emerging technology, the main focus of the literature is on detection and classification tasks, thus a subset of the tracking pipeline.
Some works explore both neural architectural optimisation and inference optimisation employing custom compilers and operations. In \cite{lin2020mcunet}, an MCU-oriented NAS (TinyNAS) is combined with a lightweight inference engine (TinyEngine), enabling ImageNet-scale inference on MCU.
Industry-oriented tools as STM X-Cube-AI can be exploited to implement artificial neural networks on MCUs \cite{s20092638}, though compromising the inference performance with respect to manual network implementation via CMSIS-NN \cite{lai2018cmsis}.
On the other end, some works \cite{roma} exploit extreme parameter reduction using XNOR Networks. However, the performance of those approaches, and in general of Binary Neural Networks (BNNs) is notably lower than the one achieved by classical CNNs \cite{bulat2019xnor} thus a lot of application scenarios are not addressable with BNNs.
Another way to implement vision intelligence at the edge is by exploiting custom hardware architectures, which use parallel computing to speed up computation \cite{garofalo2020pulp, flamand2018gap}.

In this context, our work is towards the design of a novel scalable backbone, which maximise resource usage by decoupling the computational requirements of the neural network, thus not relying on custom hardware or compilers. 

In the following sections, we describe the proposed network architecture family and how it achieves good performance in object detection and tracking suitable for microcontroller-scale execution.

\section{\textit{PhiNets} Architecture}

When constraining computational cost and memory usage to fit neural networks on an MCU, scaling approaches like the ones presented in EfficientNet highlight the inefficiency of current state-of-the-art architectures. This is proved by the significant performance drop-offs on computer vision tasks when these networks are constrained to lower powered devices, as demonstrated by the authors in \cite{howard2017mobilenets} and also confirmed empirically in Fig. \ref{fig:detresults_params}.

\begin{figure}[h]
  \centering
  \includegraphics[width=10cm]{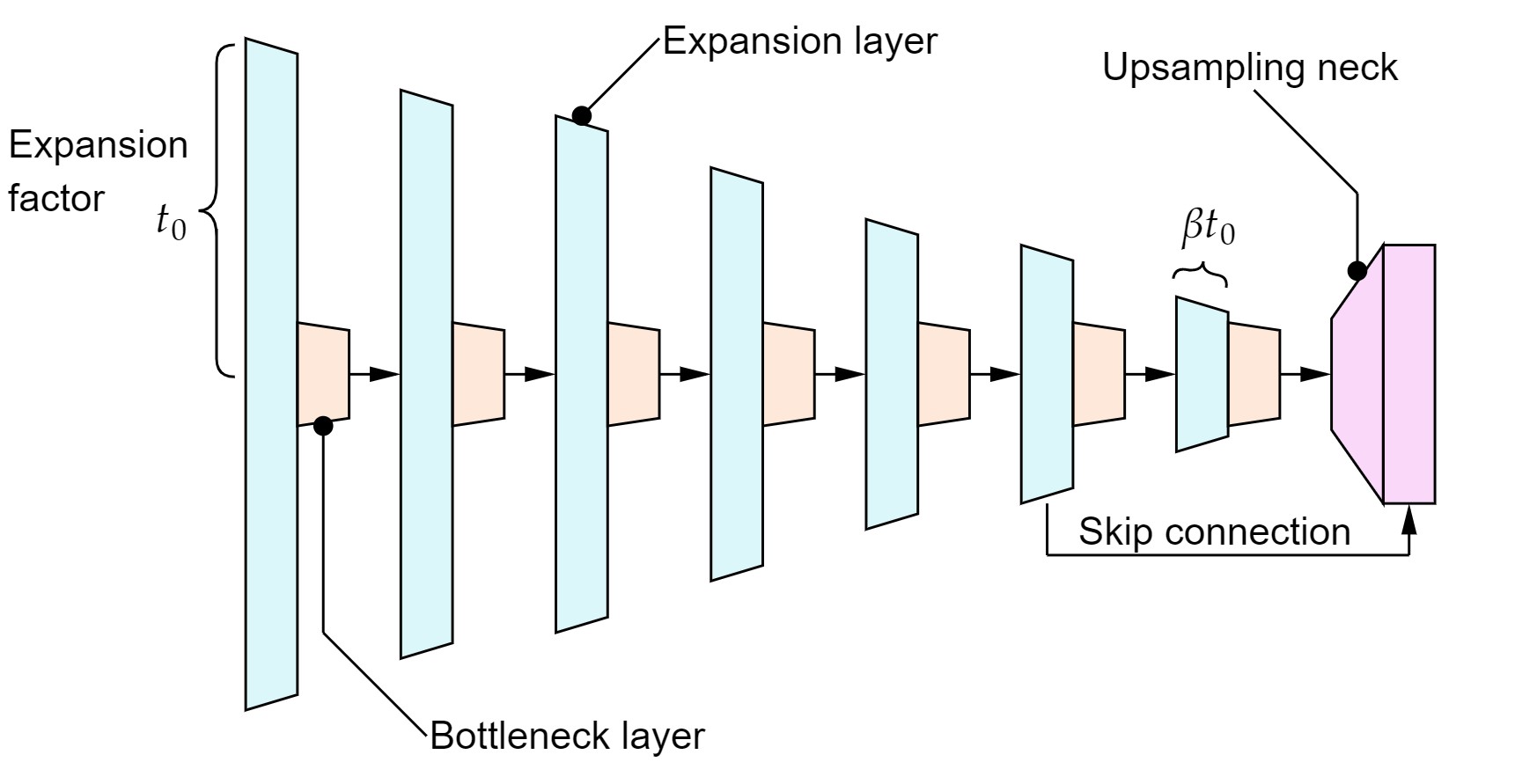}
  \caption{An overview of the \textit{PhiNets} family network architecture. The proposed architecture scales with respect to the expansion factor $t_0$, number of convolutional blocks, shape factor $\beta$ and width multiplier as explained in Sec \ref{scaling}.}
  \label{fig:phinet_overview}
\end{figure}

In this work, we present \textit{PhiNets}, an efficient neural network family developed for MCU inference. \textit{PhiNets} aim at solving the main drawbacks of current state-of-the-art scalable backbones for image processing at the edge. They are a family of networks optimized for inference within the one to ten million MACCs range, tuned to minimize the performance's drop while scaling with the architecture specifications.


In the following subsections, we will introduce the main network building blocks (\ref{bblocks}), the main constraints when it comes to MCU inference (\ref{has}) and how \textit{PhiNets} solve this problem (\ref{scaling}). In the end, will we present the detection and tracking pipelines selected for the benchmarking (\ref{sec:dettrack}).

\subsection{Network building blocks}
\label{bblocks}
As it is common in parameter-efficient architectures, such as \cite{sandler2018mobilenetv2, tan2021efficientnetv2, he2016deep}, the network is composed of a sequence of inverted residual blocks (by default, there are seven blocks for our architecture), each followed by a swish activation function. Fig. \ref{fig:phinet_overview} shows the network architecture overview.

\begin{figure}[h]
  \centering
  \includegraphics[width=14cm]{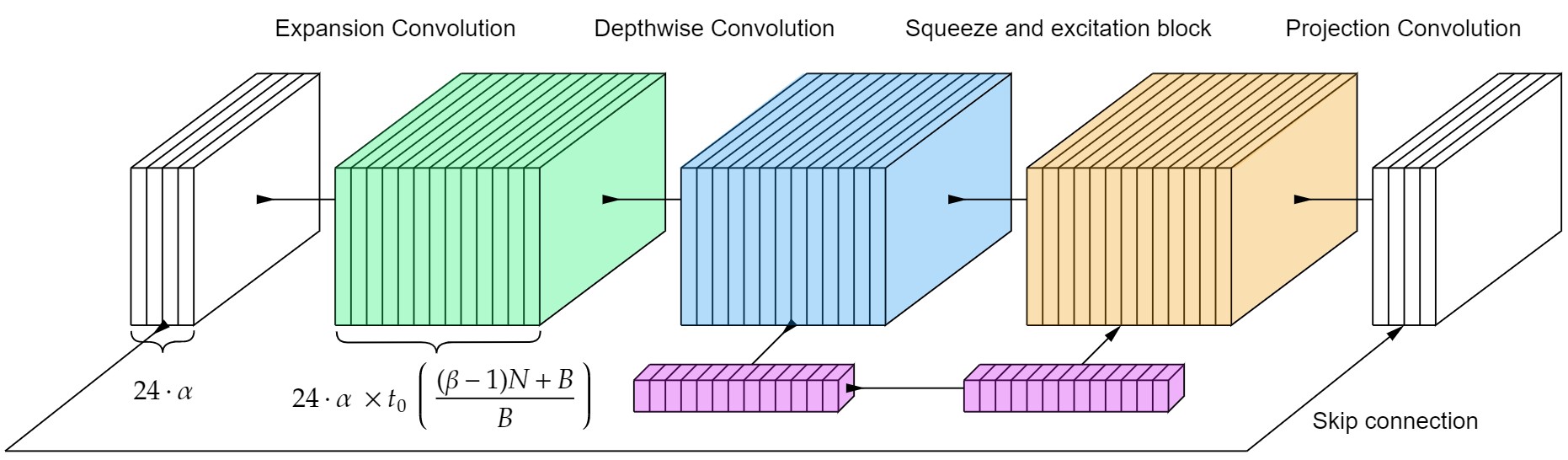}
  \caption{An overview of the \textit{PhiNets} convolutional block structure. First, the number of channels is increased with a pointwise convolution, followed by a depthwise convolution and SE block. Finally, a second pointwise convolution connects to the low dimensionality bottleneck block.}
  \label{fig:convblock_structure}
\end{figure}

As illustrated in Fig. \ref{fig:convblock_structure}, the number of filters in the first bottleneck layer is $24 \alpha$, where $\alpha$ is a hyperparameter that works similarly to how it does in the MobilenetV2 architecture, while the multiplication factor gets doubled every time the feature map is downsampled in the network. Squeeze-and-Excitation blocks \cite{hu2018squeeze} are inserted after each convolutional block and skip connections are used between the same-resolution bottleneck layers as in MobileNetV2 \cite{sandler2018mobilenetv2}.
The expansion factor used in the inverted residual block of index $N$, where N equals 0 for the first layer, is determined by two hyperparameters, i.e. $t_0$ (\textit{base expansion factor}) and $\beta$ (\textit{shape hyperparameter}), as 
\begin{equation}
    t=t_0 \bigg( \frac{(\beta-1)N+B}{B} \bigg)
\end{equation}
Five strided convolutions are used for down-sampling the feature maps through the network with a spatial resolution reduction of a factor of $32\times$ between input and output tensors.

As the network has been primarily developed for object detection tasks, we maximized the receptive field of each element in the output grid, also considering the resolution of the output tensor. Reducing the latter too much would affect the spatial information flow through the network and significantly lower the performance \cite{bochkovskiy2020yolov4}. To tackle this issue, we placed a neck composed of a $2\times$ up-sampling layer and a skip connection to the latest convolutional block of the exact resolution after the sequence of convolutional blocks.
Our approach proved to be a computationally efficient method of optimizing the trade-off between high output resolution and output grid receptive field without negatively impacting the computational requirements of the network.

\subsection{Hardware-constrained and hardware-aware scaling}
\label{has}
When bringing CNNs to edge devices, the general approach consists of (i) architectural space exploration to identify the best performing networks for a given task (ii) selection of the best network fitting in the most restrictive resource constraints. This approach is known as hardware-constrained network architecture search and is the most often used technique for architecture search in constrained devices. While this helps optimise a network for a very resource-limited device such as a microcontroller, this approach results in a network that provides sub-optimal resource utilization of the platform, as only the most stringent requirement is met. But, when working with a resource constrained platform, we usually need to face three different constraints:
\begin{itemize}
    \item the number of operations (MACC) required for network inference. High-performance microcontrollers, coupled with efficient inferencing frameworks, can achieve tens to thousands of MACC per second. For real-time video applications, this means that a top-of-the-line MCU can run a $10MMACC$ detection network at $\approx 10Hz$;
    \item the dynamic memory (working memory - WM). When executing a network's computational graph, at each point, the CPU must compute a matrix multiplication between the output of the previous layer or the input of the network and the following channel filter matrix. The RAM usage is determined by the size of these tensors, plus the memory required to keep the tensors for skip connections for later usage;
    \item the static memory (parameter memory - PM). As FLASH memory is the most expensive part of the microcontroller's die both in terms of cost and area, this can usually contain $100KB$ to $1MB$ of data. Assuming that all network parameters get quantized to 8bit integers, a maximum of $100K$ to $1M$ parameters can be used, based on the selected platform.
\end{itemize}
Our work proposes an optimized architecture family, \textit{PhiNets}, which inverts the hardware network architecture search paradigm, using a \textit{hardware-aware} network scaling pipeline. Resource constraints can be met with minimal performance loss by varying different sets of hyperparameters. Moreover, eventual performance bottlenecks can be easily identified thanks to the structure of the network.

\subsection{Meeting the requirements: decoupling MACC and memory usage efficiently}
\label{scaling}
Resource usage for the three main hardware constraints of the network can be optimized in a decoupled way, i.e., using different hyperparameter combinations to meet different resource constraints and achieve optimal use of the available hardware. This optimization will allow for superior performance with respect to networks generated using hardware-constrained scaling techniques. The following sections will highlight how the network parameters are connected to the resource requirements for \textit{PhiNets}.

\subsubsection{Number of operations}
The number of operations (MACC) for the baseline network depends on network input resolution $w\times h$, on the network width, which scales quadratically with the parameter $\alpha$, and on the network depth (determined by the number of blocks $B$). These parameters can be tuned using a compound scaling methodology as proposed in \cite{tan2021efficientnetv2} to obtain the best performing network for a given operation count or can be determined by other implementation factors (e.g. the resolution of the camera used can set a fixed $w\times h$).

\begin{figure}[H]
  \centering
  \includegraphics[height=5cm]{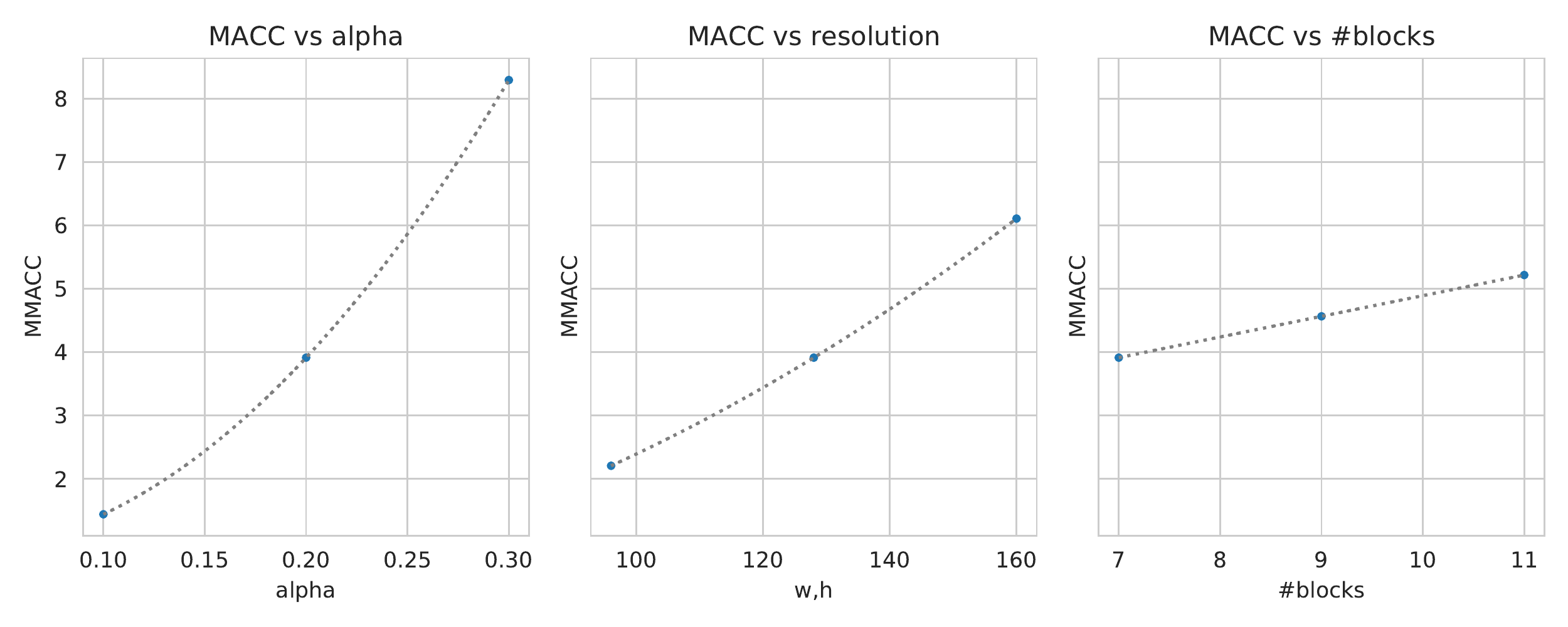}
  \caption{\textit{PhiNets} computational complexity with respect to $\alpha$, $w\times h$ and $B$. The plots highlight an exponential increase in MACC count with the number of filter and input resolution, while a linear trend with respect to the number of convolutional blocks.}
  \label{fig:macc_dependency}
\end{figure}

The parameters $w\times h$, $\alpha$ and $B$ determine the number of operations for the base network. This can be defined by real-time requirements for the system, power consumption targets, or accuracy requirements. Fig. \ref{fig:macc_dependency} shows the effects of the three parameters on the complexity of the network.

\subsubsection{Dynamic memory}
The dynamic memory will be determined by the size of the tensors in the expansion layers of the first convolutional block. The tensor size in the later layers increases linearly with the network's depth and decreases quadratically with resolution. Moreover, no tensors need to be kept in memory for the first block as there are no previous layers with residual connections. Varying the base expansion factor $t_0$  scales the dynamic memory required by the network linearly. Note that it is recommended to keep this parameter between $2$ and $8$, using by default $6$ for networks larger than $5MMACC$ and $5$ for networks smaller than that. Fig. \ref{fig:ram_flash_dependency} (left) shows the linear effects of the expansion factor on the working memory requirements of the network.

\subsubsection{Parameter memory}
\begin{figure}[H]
  \centering
  \includegraphics[height=5cm]{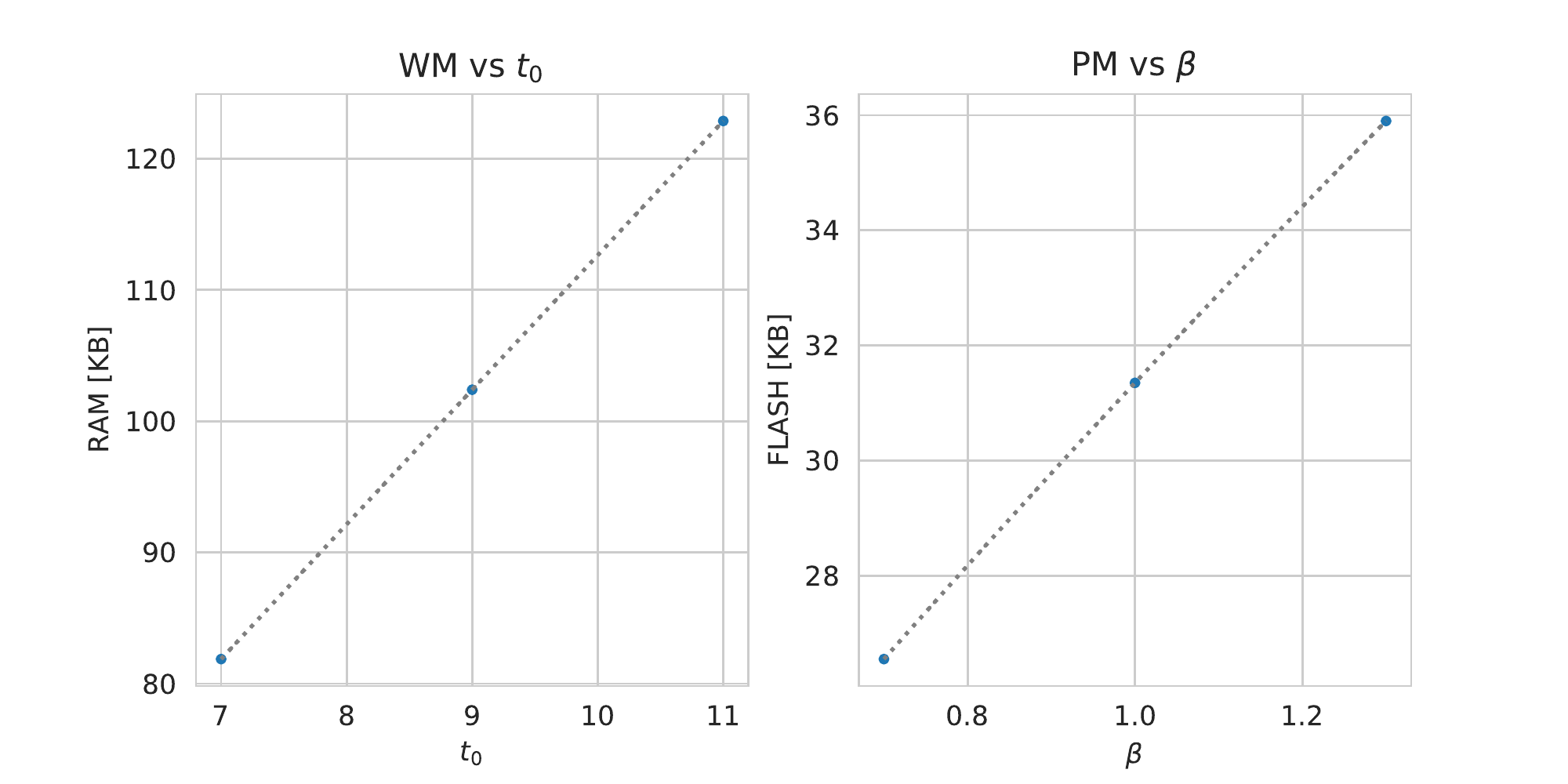}
  \caption{ Left: \textit{PhiNets} RAM requirements from a default network by varying $t_0$. Right: \textit{PhiNets} FLASH requirements from a default network by varying $\beta$. As shown, the trend is linear in both cases.}
  \label{fig:ram_flash_dependency}
\end{figure}

The parameter memory is determined by the convolutional kernels in the network. In particular, it is determined by the size of the kernels used in the expansion layers of the later network blocks, where the tensors usually have a low spatial resolution, but a high number of filters is used. Given how the expansion factor of later layers is related to $\beta$, the parameter memory required by the network varies with this parameter. In particular, the relationship between the number of parameters in the network and the shape hyperparameter is modelled as $\#Params \approx C\cdot \frac{1}{2}(1+\beta)$ with $C$ number of parameters of the base network ($\beta=1$). The Fig. \ref{fig:ram_flash_dependency} (right) shows the effects of the shape factor on the number of parameters of the network.

\subsubsection{Network tuning strategy}

While tuning the network, it is recommended to optimize the number of operations of the base network firstly, then RAM and FLASH usage. In particular, a general network-tuning approach can be as follows:
\begin{enumerate}
    \item Estimation of the available time for an inference operation and estimate of the corresponding MACC count given the platform performance;
    \item Selection of the hyperparameters $w\times h$, $\alpha$ and $B$, based either on a compound scaling technique, from a default architecture, or with a network architecture search algorithm, to achieve the correct number of operations;
    \item Tuning of the $t_0$ hyperparameter knowing input resolution and available WM from the size of input and output tensors for the first convolutional block (assuming network weights and activations get quantized to 8bit integers)
\begin{equation*}
    WM\approx \bigg(\frac{w}{2}\times\frac{h}{2}\cdot 24\alpha + \frac{w}{4}\times\frac{h}{4}\cdot 24\alpha \bigg) t_0
\end{equation*}
    For example, going from $t_0=5$ to $t_0=4$ decreases the needed RAM for the network by $20\%$;
    \item Tuning of the $\beta$ hyperparameter to achieve the desired number of parameters. Knowing the number of parameters of the starting architecture $P_0$ and the target $P_{target}$, we can obtain $\beta$ from:
\begin{equation*}
    \beta \approx 2\frac{P_{target}}{P_0}-1
\end{equation*}
    For example, going from $\beta=1$ to $\beta=0.6$ decreases the number of parameters by $20\%$;
\end{enumerate}
The optimization procedure can be repeated multiple times if higher precision is required, as varying $t_0$ and $\beta$ has second-order effects on the number of operations.


\subsection{Detection and tracking}
\label{sec:dettrack}
Nowadays, there are many alternatives for both detection \cite{he2017mask, bochkovskiy2020yolov4, liu2016ssd} and tracking \cite{zhang2020fairmot, bewley2016simple, wojke2017simple}. We took into account the algorithms considering the trade-off between computational cost and performance. For the object detection task, we used the YOLOv2 \cite{redmon2017yolo9000} detection head working at a single scale, as this requires only a single convolutional layer for bounding box prediction and class identification of all objects in the frame, leading to minimal operation count networks. Choosing YoloV2 also helps in embedded processing since the computational complexity is affected mainly by the \textit{PhiNet} architecture and does not directly depend on the number of objects in the frame.

For the tracker, we used a tracking-by-detection pipeline based on the proposed object detector. Different alternatives like SORT, DeepSORT and IoU association can be considered to work with our architecture. All of them have a computational cost that scales linearly with the number of objects to be tracked, thus implying a limit in the number of possibly tracked objects in resource-constrained platforms targeting real-time applications.
Moreover, while IoU is the shallowest concerning the operation count, it has many ID switches, thus performing worse than SORT and DeepSORT, considering this critical metric in low-fps applications, where an object might be in the scene a total of 10-20 frames. Between DeepSORT and SORT, we selected SORT since having the embedding extractor as in the DeepSORT architecture implies using a lower complexity object detector. In tracking-by-detection pipelines, it is crucial to have good detection performance since it directly impacts tracking performance. Thus, we are interested in having the best performing detector possible since it will help the shallower tracker achieve the same MOTA score as the more complex one, working with a worse detector.

\section{Results}

Since, as already described, we used a tracking-by-detection pipeline, and the tracking performance is highly dependent on the detections, we decided to split the benchmarking in detection performance, tracking performance, and power consumption to show how each component of our pipeline was performing.

\subsection{Baseline architectures}
\textit{PhiNets} baseline architectures were tested, with the parameters summarized in the Table \ref{table:params}.

\begin{table}[htbp]
\begin{tabular}{cccccccl}
\textbf{Resolution} & \textbf{$\alpha$} & \textbf{$B$} & \textbf{$\beta$} & \textbf{$t_0$} & \textbf{MACC} & \textbf{Parameters} & \textbf{Task} \\ \hline
$128 \times 128$ & $0.35$ & 7 & 1 & 6 & 9.85 M  & 61.2 K & Detection \\
$128 \times 128$ & $0.25$ & 7 & 1 & 6 & 6.08 M  & 37.9 K & Detection \\
$96 \times 96$   & $0.25$ & 7 & 1 & 5 & 3.01 M  & 31.8 K & Detection \\
$96 \times 96$   & $0.15$ & 7 & 1 & 5 & 1.23 M  & 14.3 K & Detection \\
\hline
$160 \times 160$ & $0.3$  & 7 & 1 & 5 & 10.42 M & 39.9 K & Tracking  \\
$160 \times 160$ & $0.2$  & 7 & 1 & 5 & 4.96 M  & 21.6 K & Tracking  \\
$128 \times 128$ & $0.2$  & 7 & 1 & 5 & 3.18 M  & 21.6 K & Tracking 
\end{tabular}
\caption{Parameters for generating the benchmarked \textit{PhiNets}}
\label{table:params}
\end{table}

Networks have been grouped by task, as we found that a slightly lower resolution but a higher number of filters benefited, at the same MACC count, more the detection task than the tracking one, for which a higher resolution proved to be the better choice.

\begin{figure}[H]
  \centering
  \includegraphics[height=5cm]{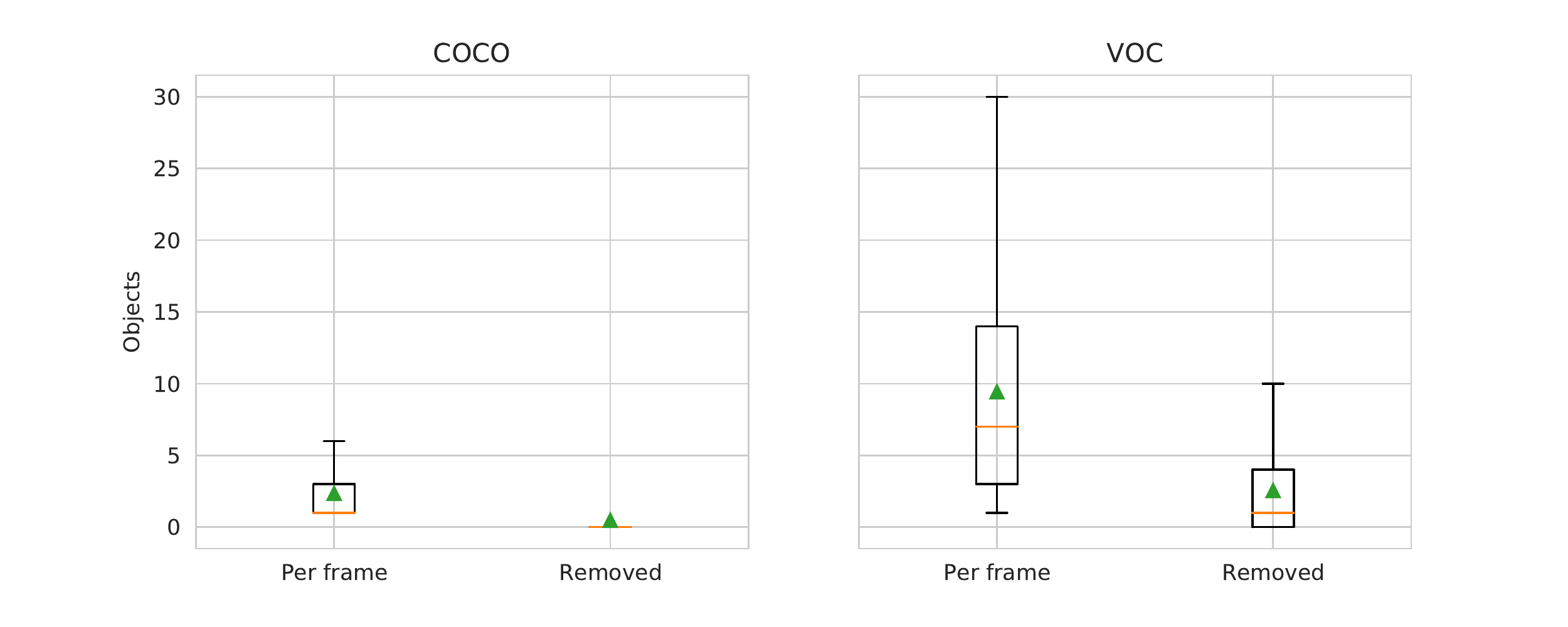}
  \caption{Number of objects removed per frame from COCO and VOC2012 datasets based on area constraint. The removal prevents having objects which are represented by only a couple of pixels on the down-sampled image (input of the backbone).}
  \label{fig:data}
\end{figure}

\subsection{Detection}
To evaluate object detection performance towards tiny multi-object tracking, we trained EfficientNets, MobileNets and \textit{PhiNets} sized between $1M$ and $10M$ MACC on a subset of the MS COCO \cite{lin2014microsoft} and VOC2012 \cite{pascal-voc-2012} object detection benchmarks.

\begin{figure}[h]
  \centering
  \includegraphics[width=14cm]{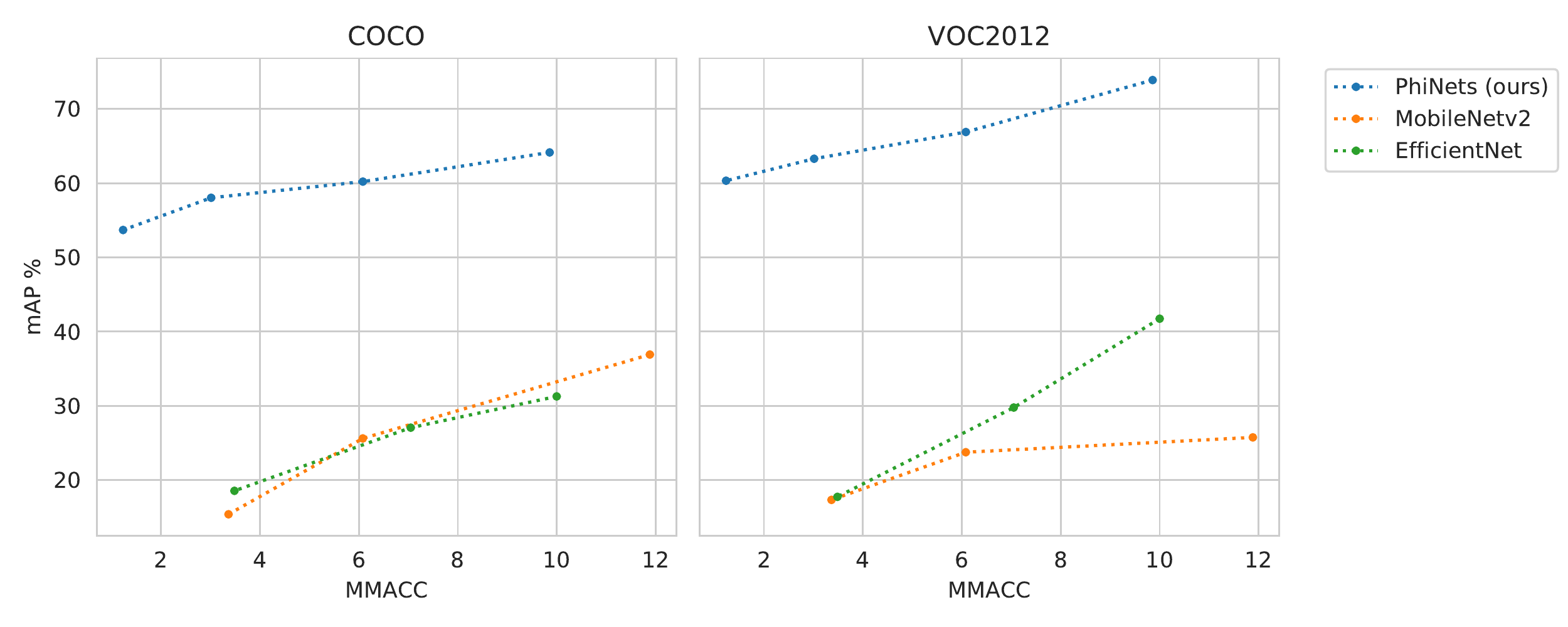}
  \caption{Comparison of scalable backbones applied before a YOLOv2 detection head for MCU-scale object detection considered in terms of  mean Average Precision (mAP) vs MACC count. The best fitting and performing models for MCU inference are the ones in the top-left area of the plot (requiring less operations with better performance).}
  \label{fig:detresults_macc_1}
\end{figure}

Given the application constraints of tiny vision, mainly regarding the input resolution, we reduced the training set by considering only the "person" class and by using only targets whose bounding box was more extensive in area than $1/64$ of the original image size. This pre-processing of the dataset is visualised in Fig. \ref{fig:data}, where we investigated the number of objects per frame in both datasets and the number of removed objects based on the above constraints.

The networks were trained for 60 (COCO) / 120 (VOC) epochs, using the Adam optimizer. The first three epochs for both datasets are used for warm-up, with a $5\times 10^{-3}$ learning rate, while for the remaining epochs, we used cosine decaying on the learning rate, starting from $1\times 10^{-2}$.

As shown in Fig. \ref{fig:detresults_macc_1}, all the \textit{PhiNets} perform better in object detection on the 1-10 MMACC range, given their advanced scalability features. In fact, \textit{PhiNet}'s performance is almost constant with respect to the number of MMACC, while EfficientNets and MobileNets have a performance drop greater that 15\% mAP in the depicted MACC range.

\begin{figure}[htbp]
  \centering
  \includegraphics[width=14cm]{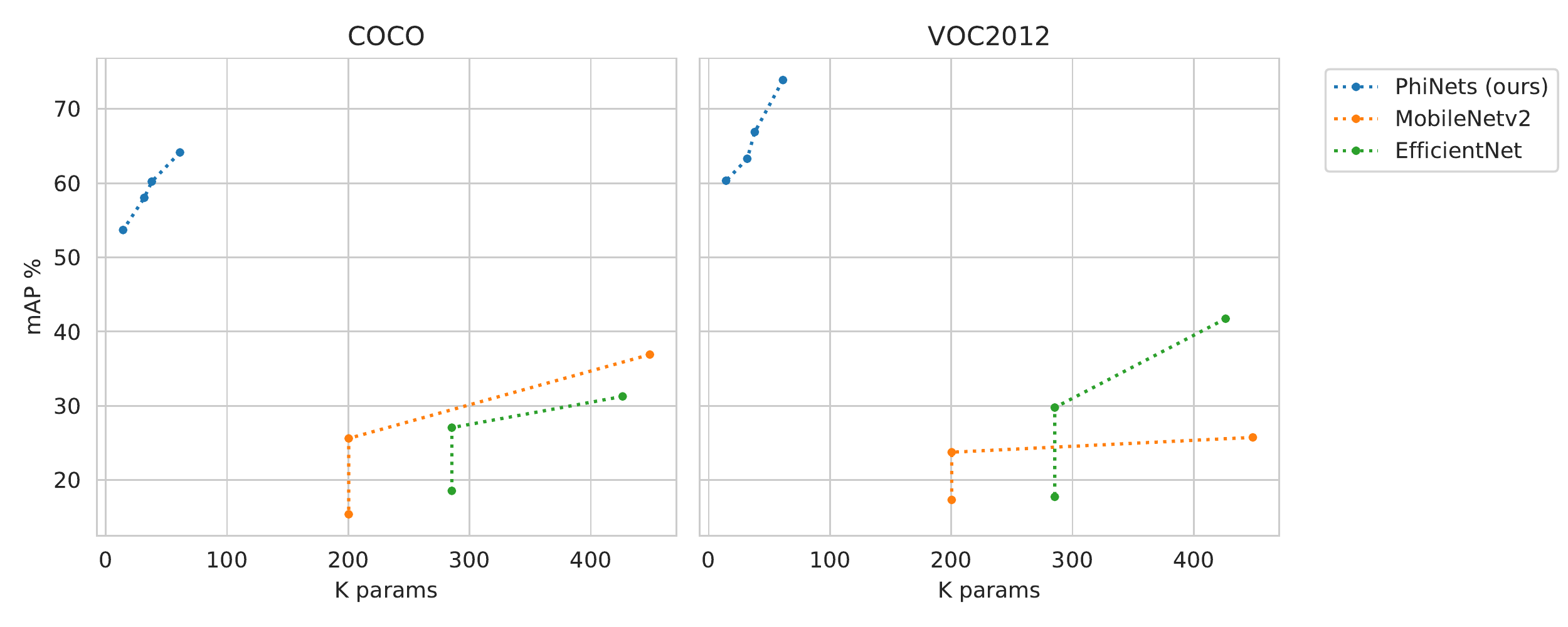}
  \caption{Comparison of scalable backbones applied before a YOLOv2 detection head for MCU-scale object detection. The plots show a vertical section when we downsized the network by lowering the input resolution, keeping the same number of filters (and thus parameters), in order to avoid drastic performance drop in EfficientNets and MobileNets. The best fitting and performing models for MCU inference are the ones in the top-left area of the plot (requiring less parameters with better performance).}
  \label{fig:detresults_params}
\end{figure}

We achieved the same behaviour also in the analysis with respect to the number of parameters, depicted in Fig. \ref{fig:detresults_params}, proving that \textit{PhiNets} are more efficient in parameter count than previous state-of-the-art architectures. In conclusion, \textit{PhiNets} set a new standard for embedded object detection on MCUs by achieving higher performance with less operations and parameters.

\subsection{Multi-Object Tracking}

After the detection, we perform multi-object tracking using SORT \cite{bewley2016simple}. We benchmarked the proposed backbones performance on the MOT15 dataset \cite{leal2015motchallenge} after augmenting the training of the detectors with 360 epochs on the benchmark data.

\begin{figure}[H]
  \centering
  \includegraphics[width=12cm]{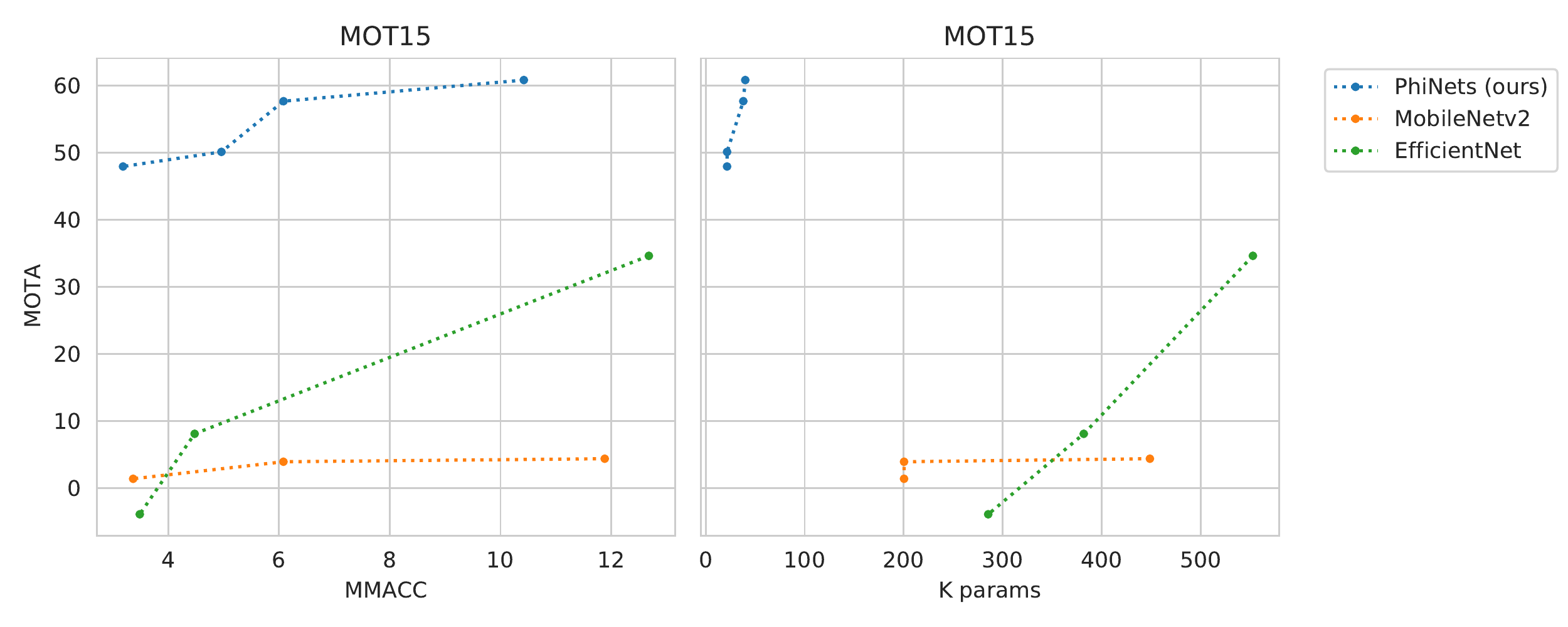}
  \caption{Results for MOT15 tracking showing the relation between Multi Object Tracking Accuracy (MOTA) score and network complexity (in terms of MACC count - left - and parameters count - right). The plots show a vertical section when we downsized the network by lowering the input resolution, keeping the same number of filters (and thus parameters), in order to avoid drastic performance drop in EfficientNets and MobileNets. The best fitting and performing models for MCU inference are the ones in the top-left area of the plot (requiring less parameters and operations with better performance).}
  \label{fig:mot_res}
\end{figure}

The relationship between detection and tracking performance in tracking-by-detection pipelines is intuitively linear. In fact, since the tracking is based on the detection IoU, a bad detector implies low tracking performance. Thus, as expected by analysing the detection results, and as we empirically proved in Fig. \ref{fig:mot_res}, \textit{PhiNets} are the best performing backbone for tracking in the 1-10MMACC range.

\subsection{Power consumption}

\begin{figure}[h]
  \centering
  \includegraphics[width=12cm]{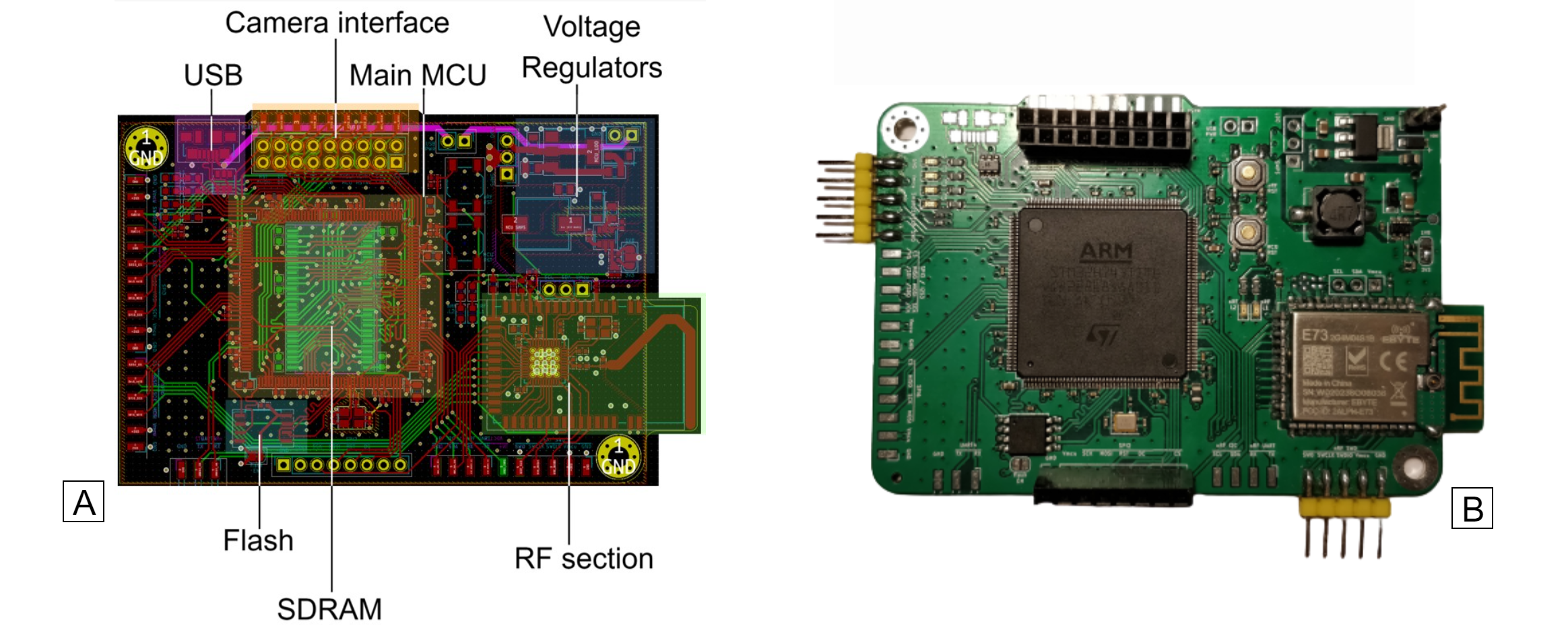}   
  \caption{The system hardware prototype realized. The board includes an \texttt{STM32H743} MCU, switch mode power regulator, a DCMI camera interface and display connection, external FLASH and DRAM and a bluetooth section (FLASH, DRAM and BLE are not needed for this application).  A: Layout of board and functional blocks.  B: Final prototype picture, with soldered components}
  \label{fig:hw_prototype}
\end{figure}

\begin{figure}[htbp]
  \centering
  \includegraphics[width=12cm]{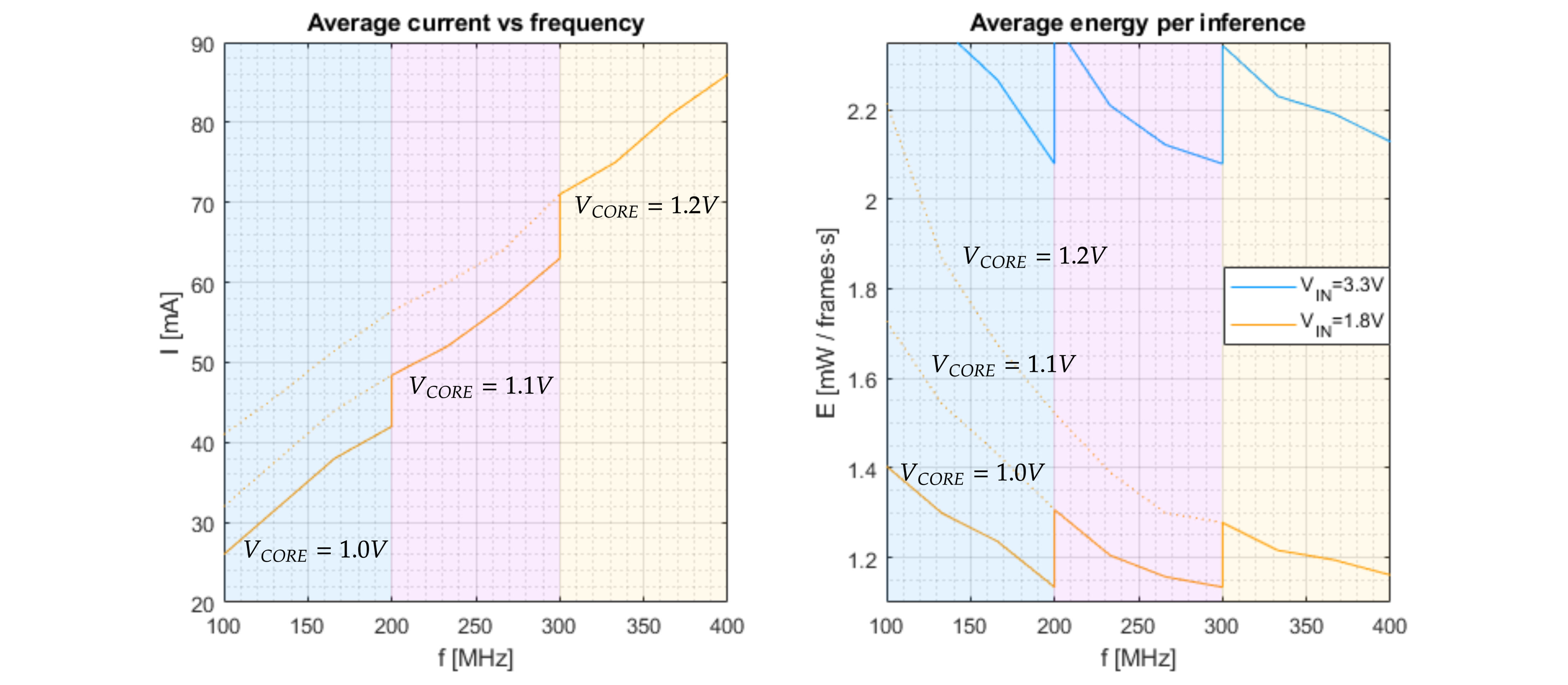}
  \caption{Trade-offs between core voltage, running frequency and power consumption for the target MCU.
  Left: current absorbed by the MCU at different running speeds when computing inference.
  Right: estimated energy per inference with the reference 1.23 MMACC network. The platform shows the minimum energy consumption running at 200 or 300MHz when powered at 1.8V. Different colors show the possible frequency ranges that can be achieved with a given core voltage. A higher core voltage (dotted line) increases power consumption without bringing additional benefits.}
  \label{fig:pwr_eff}
\end{figure}

In order to test the power consumption of the network on a off-the-shelf MCU, a prototype hardware board (shown in Figure \ref{fig:hw_prototype}) was developed, based around an \texttt{STM32H743} microcontroller, with $2MB$ of internal Flash and $1MB$ of RAM. 
The MCU was powered at $V_{dd}=1.8V$ from a switch-mode power supply, and the internal LDO powering the core was set to output $1.1V$. The microcontroller was run at a constant frequency of $300MHz$, as this allowed the best efficiency in terms of energy requirements per network inference run. This can be seen from the graphs in Fig. \ref{fig:pwr_eff}, where we analyze the effects of different running frequencies and MCU core voltages on the required current (and, consequently, energy consumption).

The energy required by the platform for an inference pass was estimated by sampling the current through a shunt resistor at the input of the hardware prototype. Empirically, we sampled the current $I(\tau)$ every $t_s=10\mu s$, and computed the energy required for inference using the relationship  $E=V_{dd} \sum_\tau I(\tau) t_s$.

\begin{figure}[htbp]
  \centering
  \includegraphics[height=5cm]{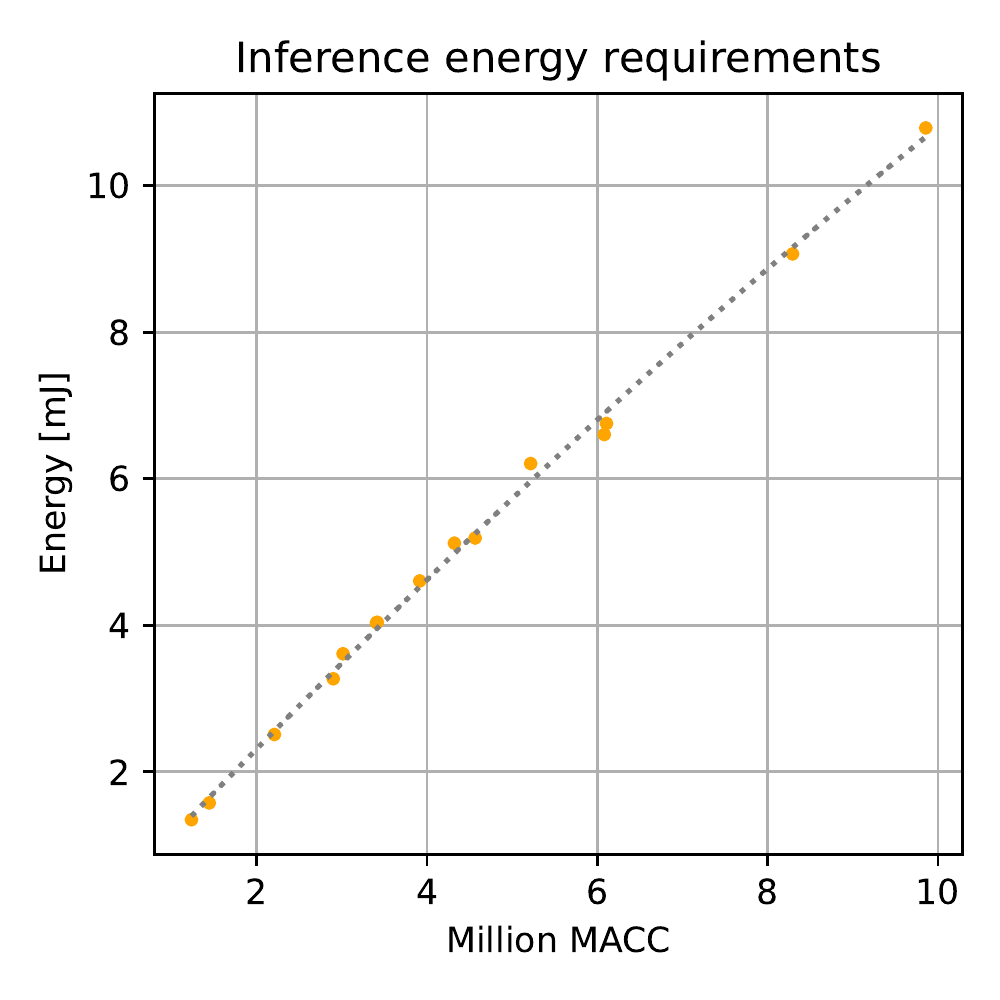}
  \caption{Linear relationship between MACC and energy requirements for the realized hardware prototype. As energy required is, in a first approximation, directly dependent on network complexity, we can interpolate data points for different networks to obtain an estimation for all possible working points of the platform (Fig. \ref{fig:average_power_fps})}
  \label{fig:Inference_energy}
\end{figure}

We investigated the relationship between computational complexity and energy required per inference in Fig. \ref{fig:Inference_energy}. After interpolating the data, we found that the relationship between MACC count and energy requirements is, in a first approximation, linear with a coefficient of around 1.2 mJ/MMACC. This, can be exploited to estimate the power consumption at different frame rates, as depicted in Fig. \ref{fig:Inference_energy}. Estimated power is computed using the data from the microcontroller's datasheet for the current in low power (between frames) and the data obtained above for the run mode current, at a certain inference time $t_L=\frac{1}{fps}$

This choice of hardware and software allowed for a state of the art power consumption of under $1.3mJ$ for the $1.2MMACC$ \textit{PhiNet} ($53.7$ / $60.3$ mAP on a subset of the COCO/VOC2012 datasets) and $11.8mJ$ for the $9.8MMACC$ \textit{PhiNet} ($64.1$ / $73.9$ mAP on COCO/VOC2012, $60.8$ MOTA on MOT15).

\begin{figure}[ht]
  \centering
  \includegraphics[height=8cm]{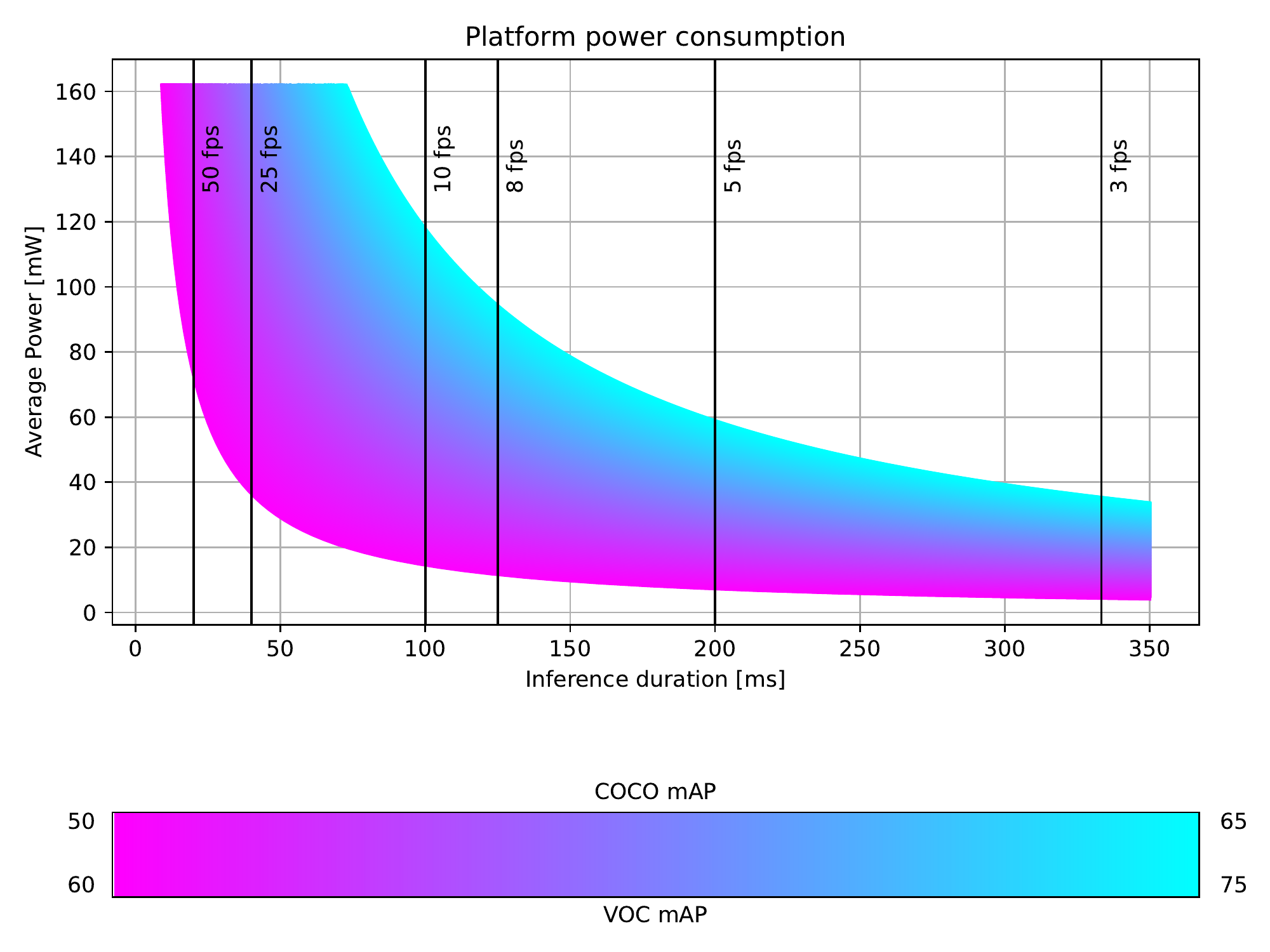}
  \caption{Possible working points of the platform, with respect to latency, performance and power consumption. Networks with higher precision are in cyan, networks with higher performance and lower power requirements in purple. Smaller networks can achieve either very low power consumption figures, or very high speeds, with, in specific fields, small accuracy losses with respect to larger ones. A usage example will be presented in section \ref{sec:example}}
  \label{fig:average_power_fps}
\end{figure}

Figure \ref{fig:average_power_fps} shows the working points that can be reached with the proposed hardware and software with respect to performance, power consumption, and inference speed. The platform is capable of running object detection at over $50$ fps with the proposed hardware, at a power consumption from $1.3mW/fps$ (for networks achieving $53.7$ / $60.3$ mAP on the selected subsets of COCO/VOC datasets) to $11.8mW/fps$ ($64.1$ / $73.9$ mAP on COCO/VOC), or, in other words, $10fps$ tracking at $13mW$ to $118mW$, depending on the performance required by the specific application.

\subsection{Case study: monitoring application}
\label{sec:example}
\begin{figure}[h]
  \centering
  \includegraphics[scale=.25]{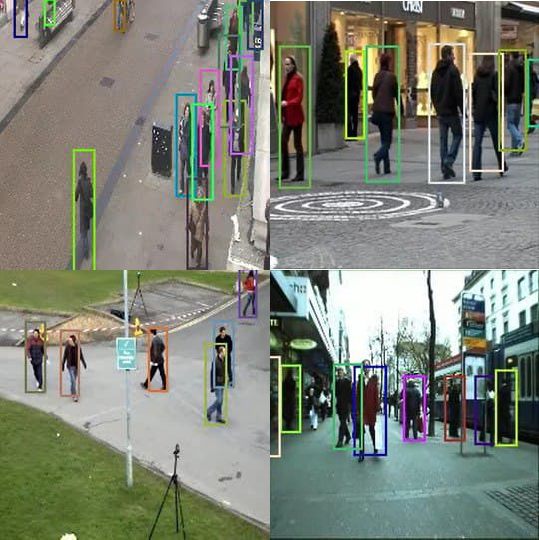}
  \caption{Visual example of detection and tracking pipeline's output.}
  \label{fig:detresults_macc}
\end{figure}

A brief case study is here presented considering a scenario akin the ones presented in the sequences \texttt{TownCentre}  (top-left in Fig \ref{fig:detresults_macc}) and \texttt{PETS09} (bottom-left in Fig \ref{fig:detresults_macc}), where a camera is mounted on an elevated position, overseeing a walkway.

Since the camera is mounted on an elevated point, there are no significant variations in the size and pose of the targets, thus allowing good tracking performance using smaller networks, which require lower power consumption. For example, the resulting MOTA score on \texttt{PETS09} is $62.86$ for the smallest \textit{PhiNet} and $68.51$ for the largest one tested. Given the large area seen by the camera, we can use low frame rates since targets will require multiple seconds to cross from one side of the image to the other. We can estimate the required energy usage for always-on tracking from Fig \ref{fig:average_power_fps}.

Given the target experimental results, we can use the smallest tracking network presented in Table \ref{table:params}, at $3.18 MMACC$ complexity. Fig. \ref{fig:Inference_energy} shows the measured energy consumption per inference at $3.21 mJ$, or $16 mW$ at the chosen framerate. A similar result can be extrapolated from the plot in Fig. \ref{fig:average_power_fps}, knowing the target frame rates and noting that the network we are using is on the left side of the mAP bar.

Such low energy requirements are ideal for always-on IoT nodes, and can run from a tiny solar panel. We tested the endnode using a $9cm \times 5cm$ monocrystalline solar panel, producing a peak power of around $913mW$. Assuming a conversion efficiency of $85\%$, a single hour under the sun could allow the system to run for two days uninterrupted.


\section{Conclusion}

In this paper we presented a new scalable backbone for low computational complexity image processing based on inverted residual blocks. The backbone was benchmarked on detection and tracking tasks with state-of-the-art results. We achieved ~20\% higher mAP with respect to EfficientNets and MobileNets on the COCO and VOC2012 detection benchmarks for the same computational complexity and ~80\% less parameters. Moreover, we achieved significantly higher MOTA score with the same compression factor.
The main asset of \textit{PhiNets} is that our model outperforms the hardware-constrained scaling by disjointly optimizing MACC, working memory and parameter memory as per the hardware-aware architecture scaling paradigm.

In conclusion, we proved that our approach successfully runs on a IoT endnode with a power consumption that allows the node to be powered using a solar panel.

\begin{acks}
  This work has been supported by the European Union’s Horizon 2020 research and innovation program under grant agreement no. 957337 (MARVEL project). This paper reflects only the authors' views and the European Commission cannot be held responsible for any use which may be made of the information contained therein.
\end{acks}

\bibliographystyle{ACM-Reference-Format}
\bibliography{bib}

\end{document}